\documentclass{article}
\usepackage{graphicx} 
\usepackage{subcaption}
\usepackage{float}
\usepackage{amssymb}
\usepackage{amsmath}
\title{FUnc-SNE: A flexible, Fast, and Unconstrained algorithm for neighbour embeddings.}
\author{Pierre Lambert, Edouard Couplet, Michel Verleysen, and John Aldo Lee}
\date{December 2024}

\begin{document}

\maketitle


\begin{abstract}

\textbf{PREPRINT SUBMITTED TO NEUROCOMPUTING}

Neighbour embeddings (NE) allow the representation of high dimensional datasets into lower dimensional spaces and are often used in data visualisation. In practice, accelerated approximations are employed to handle very large datasets. Accelerating NE is challenging, and two main directions have been explored: very coarse approximations based on negative sampling (as in UMAP) achieve high effective speed but may lack quality in the extracted structures; less coarse approximations, as used in FI$t$-SNE or BH-$t$-SNE, offer better structure preservation at the cost of speed, while also restricting the target dimensionality to $2$ or $3$, limiting NE to visualisation. In some variants, the precision of these costlier accelerations also enables finer-grained control on the extracted structures through dedicated hyperparameters. 

This paper proposes to bridge the gab between both approaches by introducing a novel way to accelerate NE, requiring a small number of computations per iteration while maintaining good fine-grained structure preservation and flexibility through hyperparameter tuning, without limiting the dimensionality of the embedding space. The method was designed for interactive exploration of data; as such, it abandons the traditional two-phased approach of other NE methods, allowing instantaneous visual feedback when changing hyperparameters, even when these control processes happening on the high-dimensional side of the computations. Experiments using a publicly available, GPU accelerated GUI integration of the method show promising results in terms of speed, flexibility in the structures getting extracted, and show potential uses in broader machine learning contexts with minimal algorithmic modifications. Central to this algorithm is a novel approach to iterative approximate nearest neighbour search, which shows promising results compared to nearest neighbour descent.

\textbf{PREPRINT SUBMITTED TO NEUROCOMPUTING}

\end{abstract}

\begin{itemize}
    \item A novel take on \textbf{accelerating variable tailed ($t$-)SNE, embedding data points into arbitrary dimensionalities while keeping good local structure preservation}. This bridges a gab between UMAP (speed and arbitrary dimensionalities), and $t$-SNE variants (better structure preservation).
    \item In contrast to other methods, the algorithm in itself, but not the public implementation, \textbf{can naturally adapt to dynamical datasets} (receiving new points, deleting points, drifting points) with no computational overhead.
    \item A novel algorithm to iteratively refine neighbour sets, less prone to local minima than nearest-neighbour descent \cite{NNdescent}. More importantly, the iterative KNN reveals \textbf{insights on a positive feedback loop between embedding iterations and KNN iterations}, which is exploited to the benefit of both sides.
    \item Insights on how NE can be repurposed to perform broader machine learning tasks.
    \item New insights on the effect of varying tails when modelling the LD neighbourhood distributions, extending the theoretical and experimental results introduced in papers studying the question. 
    \item A GPU implementation with an accompanying GUI, allowing interactive exploration of datasets of hundreds of thousands of points directly.
\end{itemize}

\textbf{PREPRINT SUBMITTED TO NEUROCOMPUTING}

\section{Introduction}

Neighbour embeddings (NE) is a family of algorithms aiming to reduce the dimensionality of datasets non-linearly. For that purpose, NE define neighbourhoods around each point in the original high-dimensional (HD) space, and find lower-dimensional (LD) representations of the points such that the neighbourhoods around each data point in the reduced space are representative of those in the HD space, according to some given distance function. By focusing on the preservation of local structures in the data, NE can unfold manifolds and produce meaningful embeddings of potentially very high-dimensional datasets.
Therefore, NE is now a central tool in biology \cite{Tasic, viSNE, umapCML} and any domain generating HD data can benefit from NE for either visualisation or, less commonly, in algorithmic processing \cite{landfill}.

When used for visualisation, dimensionality reduction (DR) represents data in $2$ or $3$ dimensional spaces. The main purpose is then visual data exploration, which can help develop intuition on otherwise impalpable HD structures and, further, form or investigate hypotheses about data. The visual representations can also be a support to explain a result that is rooted in data.\\

To be of interest, datasets typically have some structure, usually making their intrinsic dimensionality smaller than the ambient dimensionality of the observed HD space, due to correlations or other functional dependencies. Nevertheless, most datasets express an intrinsic dimensionality that is still significantly larger than 2 or 3. As a consequence, DR for visualisation purposes can at best represent part of the structure in such data. The challenge of DR for visualisation is that the nature of the structures that will be preserved and represented somehow depends not only on data, but also on the specific method of DR, its assumptions, design choices, inductive biases, and hyperparameters.

Figure 1 illustrates these influences on the 2-dimensional embedding of a 2-dimensional 'S' shaped plane laying in a 3-dimensional space. In the top panel, the first row evaluates the pointwise correlation between the distances from each point to all other points in LD and HD, red indicating low quality. The second row in the panel evaluates the preservation of the first $\lceil K=0.05N \rceil$ neighbours around each point. In short, the colours of the first row can be interpreted as a measure of the preservation of large-scale structures, while the second row evaluates the soundness of apparent local structures. Principal component analysis (PCA) \cite{PCA} yields a projection that preserves the global 'S' shape of the data, although the width of the 'S' is not well preserved visually, causing intrusions in the LD neighbourhoods. With $t$-SNE \cite{tSNEVDM}, belonging to the family of NE, the local structures are generally well preserved, at the price of distorting larger-scale structures. A change in the perplexity hyperparameter, which determines the neighbourhood size that $t$-SNE considers, can produce noticeable effects on the embedding, tearing the manifold apart in different places. \\
In the top view, the two rightmost columns differ only in the number of points sampled in total from the S-shaped distribution; despite that, visually, the embeddings have very little in common. Unbalanced sampling can also lead to misleading embeddings, as shown in the bottom panel. There, the bottom half of the S-shaped distribution is sampled $10$ times less frequently than the top half; as a result, the NE shrinks the less sampled half and even detaches both halves of the "S" with some hyperparameter configurations. These drastic differences are mostly due to the tendency of $t$-SNE to tear manifolds in "zones of weakness", where the local density in HD is low relative to surrounding areas. These occur naturally when then data distribution is too scarcely sampled to homogeneously capture the underlying structures in HD.\\
As a consequence, when using NE for visualisation, a grid search over hyperparameters and methods does not systematically map to the extraction of the same predictable properties of data, and one can only use broad heuristics if not in the capacity to take into account the specificities of the data distribution and sampling process.

\begin{figure}[H]
    \centering
    \includegraphics[width=1.0\linewidth]{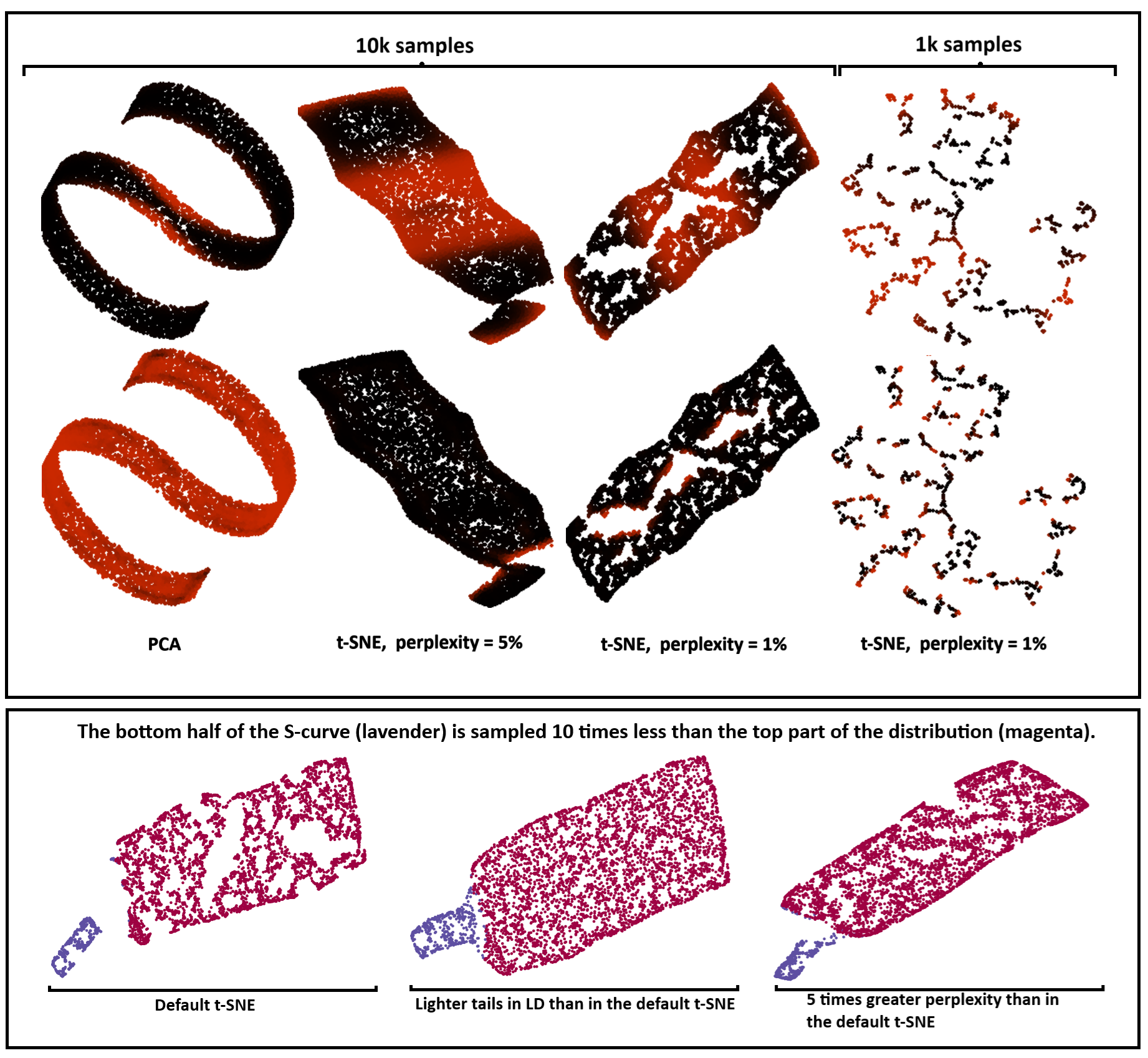}
    \caption{The impact of the DR method design choices, set of hyperparameters, and sampling of the data on the bidimensional representation of a 3-dimensional dataset.
    \textit{Top view:} the colours in the $1^{st}$ row represent how well the distances are preserved (global structures), the $2^{nd}$ row shows how well the neighbourhoods are preserved for each point (local structures); red means low quality. 
    \textit{Bottom view:} the colours label the top and bottom halves of the S-curve; the bottom half is less sampled than the top half.}
    \label{fig:localErr}
\end{figure}

The high intrinsic dimensionalities of data combined with the observation that embeddings exhibit high variability depending on the data-method-hyperparameters triad makes visual data exploration a delicate art. A proficient user must not interpret what is shown as the unique true structure of the data but rather as one partly correct, yet potentially misleading representation. To explore the dataset in a principled manner, across multiple LD
representations, one must carefully balance confirmation bias (to filter out irrelevant information) with the capability to question one's assumptions about the data.\\

From these considerations, we define three high-level properties that the ideal method of NE should exhibit.
\begin{itemize}
    \item The method should yield \textit{correct} embeddings, in the sense that the embedding is a reasonable solution given the objective of the method.
    \item The method of NE should also be \textit{flexible}: the user must be able to adjust hyperparameters in order to explore data from different viewpoints.
    \item To allow for an intuitive and interactive exploration of data from those multiple viewpoints, NE must also be \textit{fast}.
\end{itemize}

This paper proposes a novel approach to optimising a modified $t$-SNE objective function. The proposed method of NE satisfies the three desired properties by increasing $t$-SNE's speed and flexibility. The method produces embeddings of quality competitive with $t$-SNE and UMAP\cite{umap}, while keeping $\mathcal{O}(N)$ time and memory complexities and a reasonable effective number of computations per iteration. Building on the intrinsic speed and parallelisability of the proposed method, its algorithmic implementation is coded to run on GPUs and it is available in a public repository (\textit{https://github.com/PierreLambert3/fast\_htSNE}), reaching speeds compatible with interactivity for datasets of hundreds of thousands of points. 
The implementation allows for instantaneous visual feedback on hyperparameter changes, even if the hyperparameter controls properties on the HD side of the method, such as the perplexity or distance metric. Any such change in regular algorithms of NE would typically trigger a new, generally long, complete pre-computation of neighbour affinities before actually resuming the embedding optimisation. In addition to speed and correctness, the proposed method offers greater flexibility than the standard $t$-SNE or UMAP algorithms, as it allows for LD similarity distributions with tails of varying weight, following the works in \cite{parametericNE, heavytailkobak}. Experiments show that this particular property can extract structure and patterns in data that would otherwise remain hidden. The method allows for embedding data points into arbitrarily large dimensionalities, allowing its use outside of the strict visualisation context. This property was until then lacking in accelerated variants of $t$-SNE, a subgenre of NE that outperform other methods like UMAP in local structure preservation, as observed in this paper's empirical assessments and in \cite{hybrid, paperCyrilOverview}. On a side note, the proposed method of NE required a novel iterative technique to refine neighbour sets. This procedure can be used in other applications and brings insights on the possible use of directed embeddings to perform fast iterative approximate neighbourhood searches in large and potentially changing datasets.\\

In short, the main contributions of this paper are:
\begin{itemize}
    \item A novel take on \textbf{accelerating variable tailed $t$-SNE, embedding data points into arbitrary dimensionalities while keeping good local structure preservation}. This bridges a gab between UMAP (speed and arbitrary dimensionalities), and $t$-SNE variants (better structure preservation).
    \item In contrast to other methods, the algorithm in itself, but not the public implementation, \textbf{can naturally adapt to dynamical datasets} (receiving new points, deleting points, drifting points) with no computational overhead.
    \item A novel algorithm to iteratively refine neighbour sets, less prone to local minima than nearest-neighbour descent \cite{NNdescent}. More importantly, the iterative KNN reveals \textbf{insights on a positive feedback loop between embedding iterations and KNN iterations}, which is exploited to the benefit of both sides.
    \item Insights on how NE can be repurposed to perform broader machine learning tasks.
    \item New insights on the effect of varying tails when modelling the LD neighbourhood distributions, extending the theoretical and experimental results introduced in papers studying the question. 
    \item A GPU implementation with an accompanying GUI, allowing interactive exploration of datasets of hundreds of thousands of points directly.
\end{itemize}

The rest of this paper is organised as follows. Section 2 presents a detailed summary of the current state of NE for visualisation, from their advent to key technical insights under the hood of $t$-SNE, which is the specific method of NE this paper builds upon. The same section also overviews the current state of the most used methods of NE, mentioning the principal directions of research and comparing them.
Section 3 describes the proposed method of NE, starting from $t$-SNE and its close variant h-$t$-SNE \cite{heavytailkobak}. 
Section 4 evaluates the proposed method across multiple criteria and datasets, comparing the method to others of the same breed. The novel iterative process to find K-nearest neighbours is also evaluated and compared to nearest-neighbour descent \cite{NNdescent}, showing promising performances. 

\textbf{PREPRINT SUBMITTED TO NEUROCOMPUTING}

\section{The state of neighbour embeddings}

The first half of the section reviews the history of NE, motivates its adoption, and describes $t$-SNE as the most popular variant of NE to date. 

The second half of this section introduces two major directions in which research and methods of NE are aimed at: speed and flexibility. 
As its most prominent challenger in popularity, UMAP is compared to $t$-SNE with regards to the three introduced desirable properties (correctness of the LD embeddings, speed, and flexibility).

\subsection{From linear projections to neighbour embeddings}

Until 2002, the use of DR for visualisation was mainly limited to PCA and multidimensional scaling (MDS) \cite{livreMDS}. PCA is a linear projection onto a reduced set of axes that best preserves the variance of data. The linear nature of the method limits its power. Instead of preserving the variance of data, MDS relies on pairwise distances that are defined and measured in the HD space, in order to embed data in a few dimensions such that the Euclidean distances in the embedding best match those measured in the HD space. Although MDS is sometimes restricted to a linear projection, it can allow for much more expressive, nonlinear embeddings. Both PCA and MDS involve features like variance and distances whose large and thus impactful values prominently reflect the global structure of data. Therefore, they tend to extract and reproduce the global structure foremost, often overlooking more local, fine-grained structures.

When data dimensionality grows, the pairwise distances display a ratio of their mean over variance that increases too, a phenomenon referred to as concentration of norms and distance \cite{concNorms}. Intuitively, norm concentration implies that points in HD datasets tend to be nearly equidistant. As a consequence, embedding HD data in LD spaces by explicitly preserving distances is as vain as squaring a circle: LD spaces are not expressive enough to accommodate the kind of distributions of distances that live in HD spaces. In the context of visualisation in 2D or 3D, norm concentration thus restricts the use of MDS to datasets with reasonably low intrinsic dimensionalities and low noise in the ambient dimensionalities. 

In 2002, Hinton et al.~introduced a new paradigm for dimensionality reduction with their work on Stochastic Neighbour Embedding (SNE) \cite{SNE}. Their method aims to produce visual embeddings that are both resilient to norm concentration and able to capture fine-grained structures in the data by defining its loss function around the principle of neighbourhood preservation. In SNE, the authors define a differentiable function to model neighbourhoods around each point in HD and LD as a normalised Gaussian function of their distance. The distance is defined to be Euclidean in the LD space, but it can be measured otherwise in the HD space. The embedding is produced by first initialising the points randomly or in an informed manner, then, the embedding is iteratively updated by minimising the Kullback-Leibler divergence between the fixed HD Gaussian neighbourhoods and their changing LD counterparts.
In fact, SNE revealed structures in HD data that had remained unseen until then. For instance, the method visually showed for the first time a clear aggregation of numbers of the same class in the MNIST dataset, without supervision \cite{SNE}.\\

The results of SNE were groundbreaking at the time of publication, although it later became obvious that modelling neighbourhoods with Gaussian functions in the LD space led to `crowding' problems, with overly cluttered visualizations \cite{tSNEVDM}. To address this issue, $t$-SNE models the LD neighbourhoods with a normalised hyperbolic function that relates to Student $t$ distributions with a single degree of freedom. This counter-intuitive mismatch between the profiles of the HD and LD neighbourhood functions induces a distance stretch and therefore broadens the gap between clusters of points in the embedding, improving cluster visualisation but also introducing an inductive bias that fragments manifolds like dry soil. Such a tearing apart can happen on different scales, depending on the data sampling and hyperparameters, as showcased in Fig.~1.\\

Figure 2 shows embeddings of the measured transcriptome of cells in rat brains \cite{Tasic}, a data set that is further studied in \cite{artOfTsne}. Grey-brown colours show non-neuron cells, hot colours indicate inhibitory neurons, colder colours represent excitatory neurons. The PCA projection reveals an aspect of the global structure, with non neural cells located far from neurons, and the neurons themselves clearly distinguished between excitatory and inhibitory types. The MDS embedding adds further details to the structures shown by PCA. In contrast, $t$-SNE and UMAP discard the largest-scale structures, as a price to pay for revealing a finer-grained hierarchy of clusters that could not be highlighted by the use of raw distances.

\begin{figure}[H]
    \centering
    \includegraphics[width=0.8\linewidth]{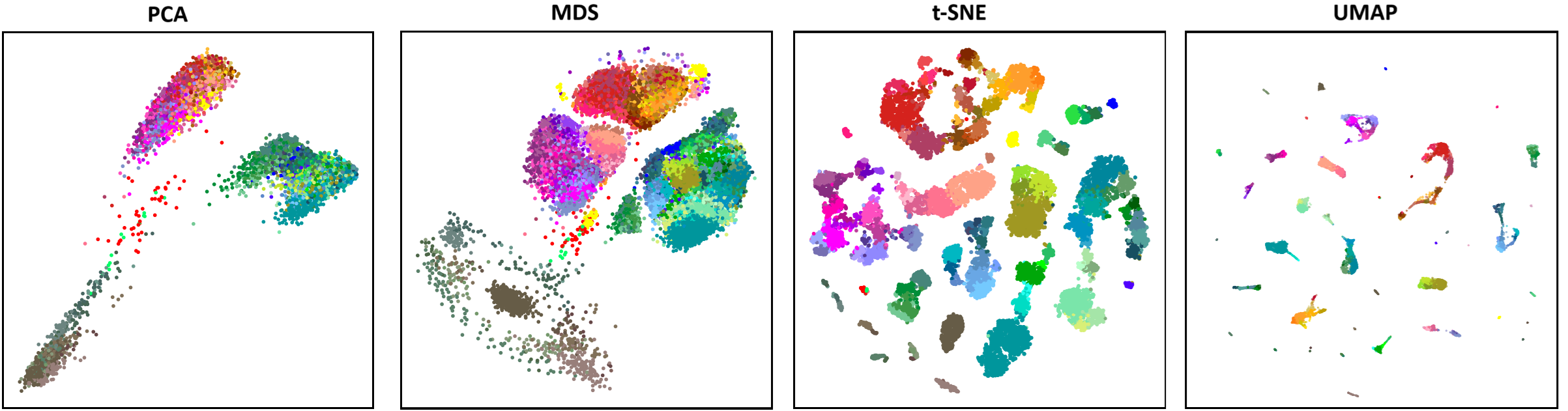}
    \caption{Visual comparison of embeddings of a single-cell transcriptomics dataset studying cells in rat brains.}
    \label{fig:ratBrain}
\end{figure}

\subsection{Student $t$-distributed stochastic neighbour embedding: $t$-SNE}

Formally, let $\mathbf{X} = [\mathbf{x}_i]_{1\leq i\leq N}$ denote the HD coordinates of the data and $\mathbf{Y} = [\mathbf{y}_i]_{1\leq i\leq N}$ be their corresponding LD representations.
Building on the fixed distances $\delta_{ij}$ measured between $\mathbf{x}_i$ and $\mathbf{x}_j$ in HD, $t$-SNE models the soft, probabilistic neighbourhoods in HD by the symmetrised pairwise normalised similarities $p_{ij}$, defined as
\begin{equation}
p_{j|i} = \frac{\exp({-\delta_{ij}^2/2\sigma_i^2})}{\sum_{k\neq i} \exp(-\delta_{ik}^2/2\sigma_i^2)} \, , \quad p_{i|i} = 0 \, , \quad \mbox{ and } p_{ij} = \frac{p_{j|i}+p_{i|j}}{2N} \enspace ,
\end{equation}
where the individual, point-specific radii $\sigma_i$ are adjusted during initialisation to comply with a user-set perplexity, this allows the HD Gaussian model of neighbours around each point to scale to the local density of the HD sample.

In the LD space, the corresponding soft probabilistic neighbourhoods are modelled through normalised similarities $q_{ij}$, defined as
\begin{equation}
q_{ij} = \frac{(1+\|\mathbf{y}_i-\mathbf{y}_j\|^2)^{-1}}{\sum_{k\neq l} (1+\|\mathbf{y}_k-\mathbf{y}_l\|^2)^{-1}} \, , \quad q_{ii} = 0 \enspace .
\end{equation}
To determine the optimal embedding $\mathbf{Y}$, $t$-SNE uses gradient descent to minimise the Kullback-Leibler divergence $\mathrm{KL}(P||Q) = \sum_{i\neq j} p_{ij} \log(p_{ij}/q_{ij})$ between the two neighbourhood distribution. The gradient for the LD coordinates gets written down as
\begin{equation}
\frac{\partial L}{\partial \mathbf{y}_i} = 4 \sum_{j\neq i} (p_{ij} - q_{ij}) (1+\|\mathbf{y}_i-\mathbf{y}_j\|^2)^{-1} (\mathbf{y}_i-\mathbf{y}_j) \enspace .
\end{equation}
At each iteration, the contribution of the $j^{th}$ point to the representation $\mathbf{y}_i$ of the $i^{th}$ point can be understood as a force aligned on $\mathbf{y}_j - \mathbf{y}_i$, which is either attractive, if the points are more similar in HD than they currently are in LD, or repulsive otherwise. \\

Most other methods of NE have their own peculiarities to model the neighbourhoods in both spaces or to define the cost function. However, they all share the common principle of constructing a representation of the dataset where embedded points are subject to pairwise attractive forces and repulsive forces, driven by the difference between sparse local affinities computed in HD, and affinities computed in LD.

\subsection{The evolution of neighbour embeddings}

The time complexity of standard $t$-SNE is quadratic in the size of data ($\mathcal{O}(N^2)$), restricting its use to datasets of a few thousands of points at most. This is problematic in some fields like the study of single-cell data in biology, where datasets can grow to hundreds of thousands or millions of points. The following paragraphs brings a quick overview of the efforts that were made to reduce the time complexity of NE.\\

Algorithmically, NE naturally decomposes into two subsequent steps: (i) all preliminary computations involving the given, fixed data, like $\delta_{ij}$, $\sigma_i$, and $p_{ij}$, and (ii) the actual optimization of $\mathbf{Y}$ and related variable quantities like $q_{ij}$. Computing the HD similarities $p_{ij}$ can be particularly slow, as it involves $\mathcal{O}(N^2)$ distances in the HD space. However, using them exclusively within Gaussian functions to characterise local neighbourhoods induces sparsity that allows accelerated methods to drastically reduce the number of considered pairwise interactions in HD, by limiting these to neighbours only. 
The neighbour sets in HD can be computed in various ways. Barnes-Hut $t$-SNE \cite{BHtSNE} uses VP trees, whereas FIt-SNE \cite{fitsne} uses multiple randomised K-D trees to quickly coalesce accurate neighbour sets. Uniform manifold approximation and projection (UMAP) \cite{umap} uses nearest-neighbour descent \cite{NNdescent}: an iterative refinement of randomly initialised neighbourhoods, where new neighbour candidates are generated from neighbours of neighbours, until convergence. LargeVis \cite{largevis} uses a combination of nearest-neighbour descent and random projections to determine the neighbour sets in HD.
While these algorithms greatly accelerate the first phase of NE, computing the HD neighbours still takes a noticeable fraction of the total runtime of NE, especially with datasets that are both large and very highly dimensional.\\

Accelerating the second algorithmic phase of NE, involving gradient descent, is not as straightforward as for the preliminary computation of HD similarities. In the LD embedding space, the similarities are generally defined using heavier tailed distributions than in HD. As a consequence, acceleration cannot rely on sparsity and short-range interactions restricted to only a few neighbours like in the HD space. Instead, fast methods of NE usually accelerate the gradient descent iterations with (locally adaptive) undersampling, like Barnes-Hut trees \cite{BHtSNE} or negative sampling \cite{largevis,umap}.\\

Negative sampling, used in UMAP and LargeVis, reduces the computational cost of the ambient field of repulsive forces by ignoring the majority of pairwise repulsive interactions. At each iteration, for each point, the global landscape of the field in LD that the point `perceives' is coarsely approximated by probing just a handful of randomly selected other points. The small resulting force is rescaled to mimic $N-1$ interactions. 
Methods of NE with negative sampling have iterations of $\mathcal{O}(N)$ time complexity, and the simplicity of the approach grants them generally higher effective speeds than other methods. However, by considering that the volume of a $n$-ball of radius $r$ increases with $r^n$, the probability of sampling a distant point is much higher than that of sampling a similar point. As a result, the local and mid-range components of the repulsive field are poorly under-estimated with negative sampling, allowing points to easily intrude, that is, to trespass unauthorized neighbourhoods. As a consequence, many seemingly close neighbours in the LD space are not actual neighbours in the HD space.
This downside is clearly apparent when evaluating the quality of embeddings with quantitative criteria such as in Fig.~\ref{fig:lesRNX} and in the numerous datasets studied in \cite{hybrid}, and in \cite{paperCyrilOverview}.\\ 

Instead of approximating the repulsive forces with unbiased random samplings, other accelerated methods, like Barnes-Hut $t$-SNE and FI$t$-SNE, model the occupancy of the LD space more faithfully by aggregating repulsive forces in a locally adaptive way. Intuitively, the idea is that if a tight cluster of points lies away enough in a certain direction, all nearly aligned repulsion vectors can be summarized into a single large vector, hardly losing any information in the process. 
Compared to negative sampling, modelling the LD space occupancy to approximate the pairwise interactions allows for more precise gradients, in particular those regarding local repulsive interactions, as supported by Fig.~\ref{fig:lesRNX} and in \cite{hybrid}. Barnes-Hut $t$-SNE and FI$t$-SNE achieve time complexities of $\mathcal{O}(N \log N)$ and $\mathcal{O}(N)$, respectively, which are both significant improvements over the original $\mathcal{O}(N^2)$, but their effective speed is generally noticeably lower than that of methods using negative sampling, due to the burden of better modelling the LD space occupancy at each iteration. Another hard limitation of such methods is that they scale exponentially with the target dimensionality, restricting their use to visualisation only.\\

\begin{figure}[H]
    \centering
    \begin{subfigure}[b]{0.9\textwidth}
        \centering
        \includegraphics[width=\textwidth]{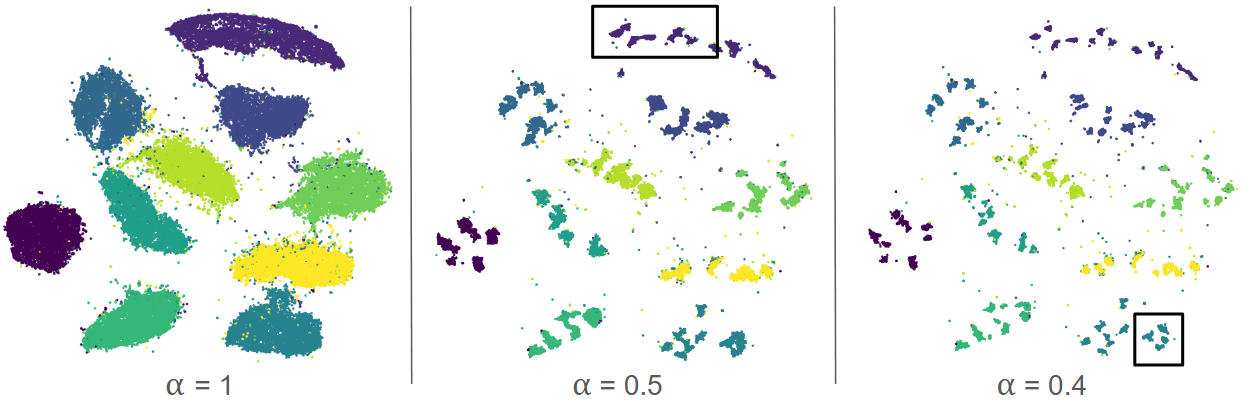}
        \caption{Representations of the MNIST dataset by the proposed method, with increasingly heavy tails on the LD similarities, from left to right.}
        \label{fig:1a}
    \end{subfigure}
    \begin{subfigure}[b]{0.45\textwidth}
        \centering
        \includegraphics[width=\textwidth]{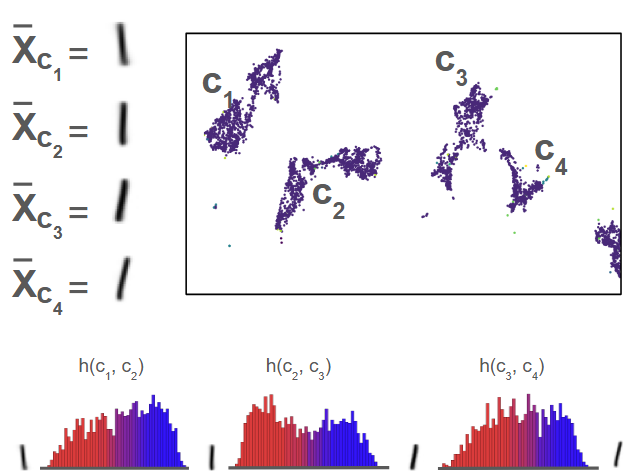}
        \caption{Subset of the "1" digits, as displayed using moderately heavier tails than $t$-SNE.}
        \label{fig:1b}
    \end{subfigure}
    \hfill
    \begin{subfigure}[b]{0.45\textwidth}
        \centering
        \includegraphics[width=\textwidth]{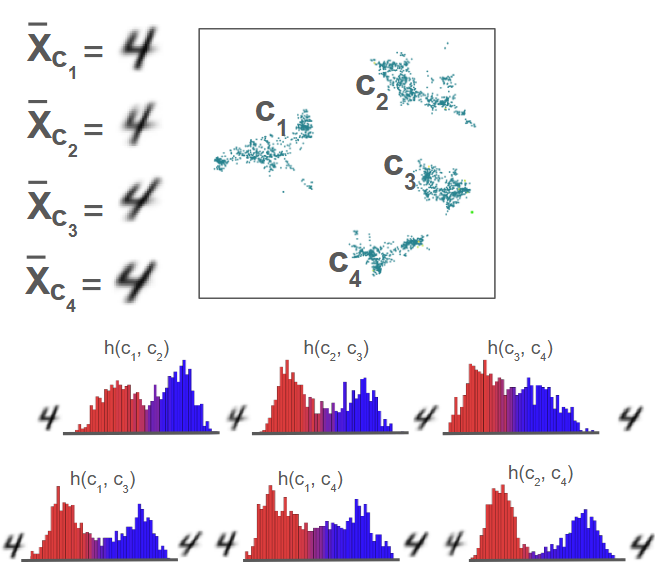}
        \caption{Subset of the "4" digits, as displayed using very heavy tails.}
        \label{fig:1c}
    \end{subfigure}
    \caption{Effect of heavier tails in the LD similarities (small $\alpha$) on the visual separation of groups of points, on the MNIST handwritten digits dataset. Embeddings produced by the proposed method.}
    \label{fig:taileffect}
\end{figure}

Beyond accelerating NE, researchers have also attempted to enhance the flexibility of methods, such that they can yield embeddings that reflect different points of view on the data, controlled by hyperparameters. In particular, researchers have investigated the effect of modifying the profile of the LD similarity distribution with respect to distance, shifting the dynamics of optimisation to reveal structures with different granularities \cite{heavytailkobak, heavytail18, NihonIchiban}. The authors show that LD similarities with heavier tails tend to produce finer-grained, more fragmented clusters, as illustrated in Fig.~\ref{fig:taileffect}.
In this figure, the handwritten MNIST dataset is represented in 3 embeddings, with varying tail heaviness in the LD similarities. Lower $\alpha$ values correspond to heavier tails, $\alpha = 1$ is equivalent to (an accelerated) $t$-SNE. Looking at the 3 whole embeddings of Fig.~\ref{fig:1a}, the large digit-specific clusters apparent using a $t$ distribution (leftmost embedding) get fragmented when increasing the `heavytailedness' of the LD similarity distribution. The views \ref{fig:1b} and \ref{fig:1c} highlight how the dynamics using heavier tails result in finer clusters. The notation $\bar{X}_{c}$ designates the mean image (in HD) of the cluster $c$ (determined in LD). In this figure, the histogram $h(c_x, c_y)$ represents the distribution of points from the set $c_x \cup c_y$ along the HD axis $(\bar{X}_{c_x} - \bar{X}_{c_y})$, after centring the set on either 
$\bar{X}_{c_x}$ or $\bar{X}_{c_y}$. The red bar colour represents $c_x$, and blue for  $c_y$.

Looking at a subset of the points with label `1' in Fig.~\ref{fig:1b}, one sees that this part of the embedding appears to be organised by the tilt angle of the `1', along the horizontal direction in the embedding. Assuming a homogeneous sampling and uniformly distributed tilt angles, one would expect these points to lie on a continuous one-dimensional manifold. The separation into multiple sub-clusters when going from $\alpha=1$ ($t$-SNE) to $\alpha=0.5$ apparent in Fig.~\ref{fig:1b} can therefore be surprising, and one can wonder whether this subdivision is truly driven by data or whether it is an unrelated artifact of the optimisation dynamics. 
The histograms reveal that the clusters get separated in LD around a dip in the probability distribution along certain directions in HD, which shows that the sub-cluster separation bears a meaning. At this point, one can only hypothesise to explain the dip in the histograms along these directions: does it stem from some unfortunate sampling? or is it a shared bias in the handwriting of humans? or perhaps the digitalisation by the camera misrepresents some angles?\\
Similar observations can be made by looking at Fig.~\ref{fig:1c}, where a subset of the digit `4' gets fragmented into 4 clusters when going from $\alpha=0.5$ to $\alpha=0.4$. Looking at the 6 possible pairs of clusters, one can reasonably conclude that, when using heavier tails, this method of NE has found directions of weakness, along which a dip in the distribution of points initiated the fragmentation of a cluster from HD to LD.\\

Exploring the landscape of embeddings produced by different LD similarity distribution shapes can greatly enhance the visualisation process, in particular when integrated to an interactive visualisation software responding visually and in real time to changes in hyperparameters or data. Such an integration allows the user to intuitively grasp the links between different conformations through the movement of the points, effectively adding another accessible dimension to the usual static embeddings. Interactivity requires methods to be very fast to (re)start and to run. Unfortunately, having tweakable LD similarity distribution shapes is not particularly adapted to the fastest methods relying only on negative sampling to estimate the effect of repulsive field on each point. This comes from the fact that the tailedness of the LD similarities has a significant role in the delicate attraction/repulsion balance across all scales \cite{attracRepulSpectrum}, including the local ones, resulting in the fine grained structures seen in (Fig.~\ref{fig:taileffect}~).

To conclude with this quick overview of NE, methods like FI$t$-SNE can precisely model the repulsive forces applied to each point, at each iteration. This accuracy makes such methods modifiable to broaden the spectrum of extracted structures. However, their accurate way to evaluate repulsive interactions in the LD space decreases their speed, reduces their interactivity, and limits the target dimensionality to 2 or 3. Other methods like UMAP run faster indeed, at the cost of embedding quality and flexibility. The next section introduces a method that bridges this gap by improving the estimation of the local repulsive forces compared to using negative sampling only, allowing greater flexibility, while still running extremely fast (smooth interactivity-enabling speeds at hundreds of thousands of points on a GPU-bearing laptop, given reasonable HD dimensionalities in the order tens or low hundreds); without constraints on the target dimensionality to 2 or 3, a property explored later in the results.

\section{A Fast, UNconstrained SNE: FUN-SNE}

Our proposal builds upon a recent generalised method of NE \cite{heavytailkobak}. The authors extend the definition of NE, with a more flexible shape for the LD similarities, which gets an additional hyperparameter $\alpha\in\mathbb{R}_{>0}$ that determines how heavy the tails are in the distribution \cite{heavytailkobak}. The LD pairwise similarities $q_{ij}$ are defined as
\begin{equation}
q_{ij} = \frac{w_{ij}}{\sum_{k\neq l} w_{kl}} 
\mbox{ with } 
q_{ii} = 0 
\mbox{ and } 
w_{ij} = \frac{1}{\left(1 + \frac{\|\mathbf{y}_i-\mathbf{y}_j\| ^2}{\alpha}\right)^{\alpha}} \enspace .
\end{equation}
The gradient then becomes
\begin{equation}
\frac{\partial L}{\partial \mathbf{y}_i} = 4\sum_j (p_{ij} - q_{ij})w_{ij}^{1/\alpha} (\mathbf{y}_i - \mathbf{y}_j) \enspace .
\end{equation}
When $\alpha$ takes values between 0 and 1, the tails of the LD similarity distribution are heavier than that of a Student $t$-distribution; $\alpha=1$ is equivalent to $t$-SNE; when increasing alpha towards infinity, the tails decrease asymptotically.\\

In order to accelerate the algorithm without constraining the LD space to $2$ or $3$ dimensions while keeping a good approximation of the local repulsive forces, this paper introduces an ensemble approach to the optimisation of the loss function. Let the twin sets of HD and LD neighbour indices for the $i^\mathrm{th}$ point be denoted $\mathfrak{N}_i^\mathrm{HD}$ and $\mathfrak{N}_i^\mathrm{LD}$, respectively, and $\frac{\partial L}{\partial \mathbf{y}_{i|j}}$ be the gradient on $\mathbf{y}_i$ resulting from interacting with the point number $j$. The cost function gradient can be re-written as
\begin{equation}
\frac{\partial L}{\partial \mathbf{y}_{i}} = \sum_{j \in {\mathfrak{N}}_i^\mathrm{HD}}\frac{\partial L}{\partial \mathbf{y}_{i|j}} +\sum_{\substack{j\in {\mathfrak{N}}_i^\mathrm{LD} \\ \And j \notin {\mathfrak{N}}_i^\mathrm{HD}}}\frac{\partial L}{\partial \mathbf{y}_{i|j}} + \sum_{\substack{j\notin {\mathfrak{N}}_i^\mathrm{LD} \\ \And j \notin {\mathfrak{N}}_i^\mathrm{HD}}}\frac{\partial L}{\partial \mathbf{y}_{i|j}} \enspace .
\end{equation}
Notice that the union of the these sets of indices encompasses all pairwise interactions for $i$, and their intersection is empty. In sufficiently large datasets, the majority of the interactions correspond to the third term of the sum, where point number $i$ interacts with points that are neither neighbours in HD, nor in LD. This part is what is approximated by negative sampling in methods like UMAP and LargeVis.\\

The first term of the sum corresponds to indices $j$ that are neighbours of $i$ in HD. This sum is the portion of the cost function responsible for the attractive pairwise interactions. All NE consider this sum somehow in their loss function. When HD neighbours get too close in the embedding, negative forces between them can be generated either as a direct consequence of the loss function, or in some methods as an additional aesthetic layer of interactions. However, commonly, a large part of the neighbours in LD are not neighbours in the HD set, which leads to the second term of the sum, a novelty of the proposed method.\\ 

The second term of the sum concerns pairwise interactions with points that are neighbours of $i$ in LD, but not in HD. This allows the proposed method to more precisely model the local LD structures around each point than methods like UMAP, which relies on negative sampling and the HD neighbours only. While the proposed method does not precisely capture the whole range of scales at which pairwise interactions occur, as FI$t$-SNE or BH $t$-SNE would, experiments show that this simple addition can bring satisfactory results with a reduced cost, assuming an efficient way to approximate neighbours in LD. This simplification also allows the proposed method to embed points into arbitrarily high dimensional spaces as UMAP can, because modelling the whole occupancy of the LD space is no longer directly required. A summary of the main directions used to approximate the repulsive field around each point is shown in Table \ref{tab:comparisonRepulsiveFields}.

\begin{table}[H]
    \centering
    \begin{tabular}{|c|c|c|c|}
        \hline
         & Close range & Medium range & Far away \\ \hline
        Negative sampling only & poor & none & correct \\ \hline
        Modelling the whole space & correct & correct & correct \\ \hline
        Proposed method & correct & none & correct \\ \hline
    \end{tabular}
    \caption{Comparison of the quality of approximations of the different components of the repulsive field around each point, for negative sampling based methods (like UMAP), methods that model the whole LD space occupancy (such as FI$t$-SNE), and the proposed approach.}
    \label{tab:comparisonRepulsiveFields}
\end{table}


This simple approach to accelerating NE while keeping local structures of quality reformulates the problem of quickly and accurately modelling the LD space, into that of quickly and accurately modelling the neighbourhoods in LD. Let $\hat{\mathfrak{N}}_i^\mathrm{HD}$ and $\hat{\mathfrak{N}}_i^\mathrm{LD}$ denote the approximate sets of neighbours around point $i$ in HD and LD, respectively. As a reminder, all other methods of NE proceed with two phases: (i) determining the exact or approximate HD KNN sets for all $i$, and (ii) using gradient descent to optimise the loss function. The proposed approach bypasses the first phase by interweaving neighbourhood discovery and gradient descent (GD), admitting that the first GD steps will yield very noisy estimates.\\

Taking inspiration from nearest-neighbour descent, the method iteratively looks at neighbours of neighbours to generate candidates for the sets $\hat{\mathfrak{N}}^\mathrm{HD}$ and $\hat{\mathfrak{N}}^\mathrm{LD}$. A peculiarity of this algorithm is that both estimated neighbours sets communicate during candidate generation: a candidate neighbour destined for $\hat{\mathfrak{N}}_i^\mathrm{HD}$ can be generated from neighbours in LD or neighbours of neighbours according to $\hat{\mathfrak{N}}^\mathrm{LD}$, and conversely. Considering that the GD steps occur at the same time, the method benefits from a convenient positive feedback loop: the better the embedding (through the gradients), the better the KNN estimation for both spaces (better candidates); closing the loop: the better the KNN sets, the better the gradients, and the better the embedding. 

\begin{figure}[H]
    \centering
    \includegraphics[width=1.0\linewidth]{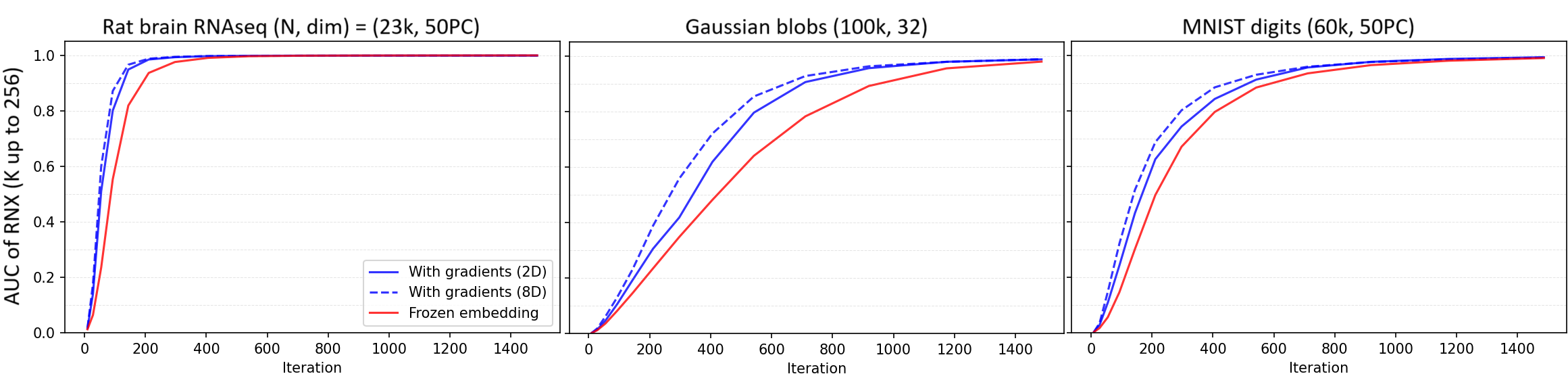}
    \caption{Evolution of the quality of the estimated KNN sets in HD throughout time, with a fixed embedding (red) or embeddings subject to optimisation (blue).}
    \label{fig:withWithoutGrad}
\end{figure}

The feedback loop is witnessed in Fig.~\ref{fig:withWithoutGrad}, where the quality of $\hat{\mathfrak{N}}_i^\mathrm{HD}$ is evaluated across iterations, using a fixed embedding (no feedback) and an embedding progressively updated through gradient descent. The quality measure is the area under the curve $R_\mathrm{NX}(K)$ \cite{RNX} (AUC of $R_\mathrm{NX}(K)$) for the first $256$ neighbours ($1\leq K\leq 256$); the estimate $\hat{\mathfrak{N}}_i^{HD}$ is compared to the exact ${\mathfrak{N}}_i^\mathrm{HD}$ computed for the purpose of this validation experiment. In short, $R_\mathrm{NX}(K)$ measures the intersection between two sets of neighbours across all scales $K$, subtracting the baseline associated with the expected rate of fortuitous neighbour matches and rescaling accordingly between $0$ (not better than purely random) and $1$ (perfect retrieval).
The AUC of the $R_\mathrm{NX}(K)$ gives a multi-scale overview of the quality of the neighbour set. 
These plots show that using the NE gradients in the LD embedding increases the speed at which the neighbour sets in HD are refined, despite not changing the KNN-finding portion of the method. Interestingly, increasing the dimensionality of the LD embedding from $2$ to $8$ seems to increase the positive effect of the feedback loop. One explanation is that the more voluminous space allows points to move more freely without negative interferences, allowing for more faithful structures in LD, which in turn helps generating better candidates for neighbour refinements in HD. A more detailed evaluation of the KNN discovery scheme is presented in the next section.

During optimisation, flags are used to keep track of points that discovered new neighbours in HD. Periodically, the flagged points have their associated adaptive bandwidth in HD ($\sigma_i$) updated using a warm restart from their previous value, for efficiency. The associated HD similarities are then updated accordingly. For datasets of reasonable size, for instance with less than $300\times10^3$ points, if running the algorithm on a modern laptop with a GPU, the quality of each KNN set in $\hat{\mathfrak{N}}^\mathrm{HD}$ quickly approaches the maximal value. To prevent superfluous computations and favour the GD part of the algorithm once the sets in $\hat{\mathfrak{N}}^\mathrm{HD}$ are close to those in ${\mathfrak{N}}^{HD}$, the refinement of $\hat{\mathfrak{N}}^\mathrm{HD}$ can be skipped. The probability of performing the refinement of $\hat{\mathfrak{N}}^\mathrm{HD}$ is arbitrarily set to $0.05 + 0.95\mathrm{E}[\frac{\mathrm{N_{new}}}{\mathrm{N}}]$, where $\mathrm{E}[\frac{\mathrm{N_{new}}}{\mathrm{N}}]$ is the smoothed fraction of points that have successfully received new neighbours in the recent past. Since the embedding is subject to a GD step at each iteration, $\hat{\mathfrak{N}}^\mathrm{LD}$ is systematically refined at each iteration. Refining the LD KNN sets is less costly than in HD, due to the lower dimensionality and because changes in $\hat{\mathfrak{N}}^\mathrm{LD}$ do not trigger the same cascade of adjustments as in HD. Naturally, iteratively updating the LD neighbours requires the learning rate to be reasonably small, to allow the sets to keep up with the changes in the LD coordinates. In practice, however, large learning rates do not seem to have a noticeable negative effect on the method, likely because allowing the use of HD neighbours to generate candidates provides a stable source of good candidates for LD neighbours that are unaffected by changes in the embedding.\\

In addition to benefitting from the aforementioned positive feedback loop, refining the KNN sets concurrently with the GD steps makes the algorithm natively adaptable to online learning, where points can be added or removed on the fly, without disturbing the flow of iterations. However, the current implementation is not designed around this use case. In the same vein, another positive aspect of abandoning the traditional two-phase approach is that the algorithm is particularly well suited to integration in interactive visualisation software. Optimisation right away after the initial memory allocations, giving the user immediate visual feedback. In practice, meaningful structures start appearing soon after launch and well before the estimated HD neighbours sets converge. Similarly, the proposed scheme allows the embedding to adapt seamlessly to hyperparameter changes in the HD side of the NE, such as the distance metric or perplexity. The current implementation is available at \textit{https://github.com/PierreLambert3/fast\_htSNE}, it comes with an interactive visualisation software from which the experiments in this work were performed.\\

In practice, having heavy tails in the LD kernels ($\alpha \ll 1.0$) tends to produce numerous visual clusters with very high intra-cluster point densities, hampering visualisation. For this reason, the proposed method keeps the attractive forces separated from the repulsive ones by distributing $p_{ij}$ and $q_{ij}$ in the gradient formula ($\frac{\partial L}{\partial \mathbf{y}_i}
= 4\sum_j \Big[ p_{ij} w_{ij}^{1/\alpha} (\mathbf{y}_i - \mathbf{y}_j)
\;-\; q_{ij} w_{ij}^{1/\alpha} (\mathbf{y}_i - \mathbf{y}_j) \Big]$); the two vectors are aggregated at the time of applying the gradients using a scaling factor as hyperparameter. This allows the user to counteract the visual collapse of clusters when using heavy kernel tails in LD by increasing the repulsive forces on the fly. The proposed method can also optionally increase the attractive forces at the start of the optimisation to facilitate the movements of points, as commonly done with early exaggeration \cite{vanDerMaaten2008} in other methods. The hyperparameter can also come handy when dealing with very large datasets, where enhancing the attractive forces is often recommended, for less crowded visualisations \cite{artOfTsne}.\\

Since the HD KNN discovery is more effective when the embedding reflects the HD data, the algorithm can benefit from a bit of help in the very first iterations to jump start the HD KNN discovery. To do so, if the algorithm is used in an interactive GUI, playing with the hyperparameters in the early iterations tends to change the embedding sufficiently to allow a good initial exploration in the HD KNN finding routine. Outside of the interactive setting, the same effect can be obtained by performing a couple of linear projections of the data towards the LD space during the first 100 or 200 iterations instead of using the NE gradients. While not strictly necessary, this can help reduce the number of iterations before stabilisation of the embedding.

A final trick worth mentioning is the recommended addition of an `implosion' button in visual software interfaces for the method. The formula of the gradient formula shows that the bigger the scale of the embedding in LD, the smaller the gradient norms. This can be problematic if visualising the data in real time through a continuous optimisation, because some combinations of attraction / repulsion ratios and LD pairwise similarity distribution shapes can lead to a seemingly ever-expanding embedding.
The software provided with the proposed method features an `implosion' button where the whole embedding coordinates are scaled down by a large factor, allowing the gradients to be significant again.\\

The method scales linearly with the number $N$ of points, and the dimensionality $M$ of the data. The linear increase in time with $M$ is particularly noticeable in the implementation accompanying this paper, because the distance computations are not parallelised, contrary to the rest of the algorithm. The time to compute the $M$-sized sum involved in most distances might be lessened by the use of parallel reduction in future implementations. When using the method for interactive data visualisation, it is recommended to first reduce the HD dimensionality of the data linearly to a manageable number of dimensions, depending on the machine and dataset size, in order to keep a high number of iterations per seconds for a smooth experience. 
In NE, it is quite common to see PCA being used for that, and keeping $50$ to $100$ components before running the NE itself \cite{artOfTsne}. When using NE for visualisation, the LD dimensionality is so low that reducing the HD dimensionality to $50$ generally does not have a noticeable impact on the visual results. Exceptions to the rule exist, when datasets already have very little redundancy extractable by linear methods, which is often the case when exploring latent spaces of deep neural networks.
It should be pointed out that the abrupt break in the quality of the repulsive field when passing from an accurate approximation for close relationships (through the approximate LD neighbourhoods) to a coarse approximation in further ranges (through negative sampling) can sometimes make the embeddings reach equilibria that are visually different from those of other methods like BH-$t$-SNE and FI$t$-SNE for identical sets of hyperparameters. In practice, changing the repulsion/attraction ratio can lead to configurations that strongly resemble those obtained by other variants of $t$-SNE like UMAP, in agreement with published results \cite{attracRepulSpectrum}.

The next section evaluates the embedding method and the KNN finding subroutine empirically, both qualitatively and quantitatively.

\section{Empirical evaluation}

Subsection 1 evaluates the method in its capability to produce embeddings for the purpose of visualisation, both qualitatively and qualitatively. Its running time is also assessed. Subsection 2 evaluates the capacity of the algorithm to model data at higher dimensionalities that 2 or 3. 

\subsection{The case of data visualisation}

\begin{figure}[H]
    \centering
    \begin{subfigure}[t]{0.49\linewidth}
        \centering
        \includegraphics[width=\linewidth]{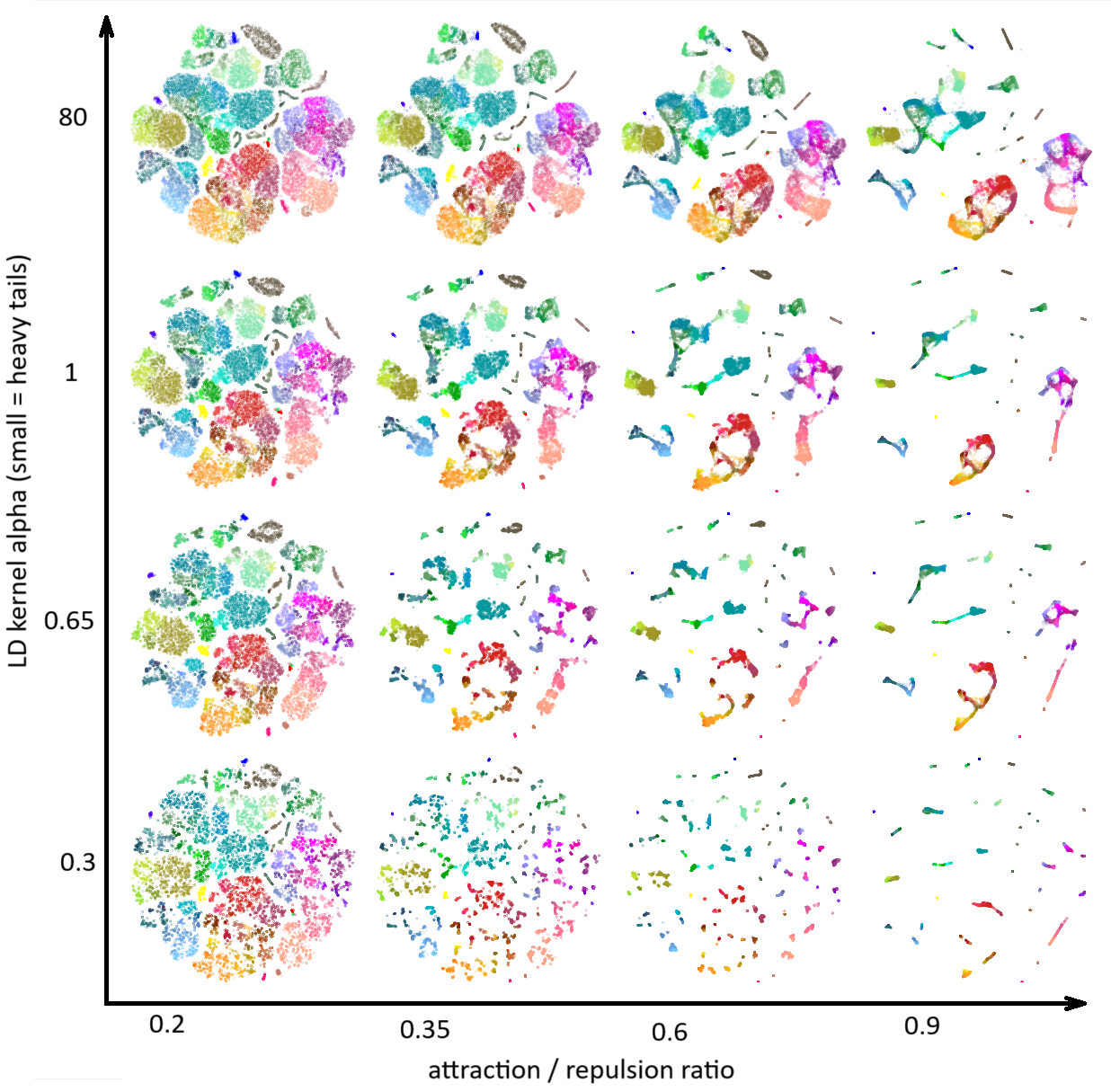}
        \caption{RNA-seq.: rat brain}
        \label{fig:qualitative_rna}
    \end{subfigure}
    \hfill
    \begin{subfigure}[t]{0.49\linewidth}
        \centering
        \includegraphics[width=\linewidth]{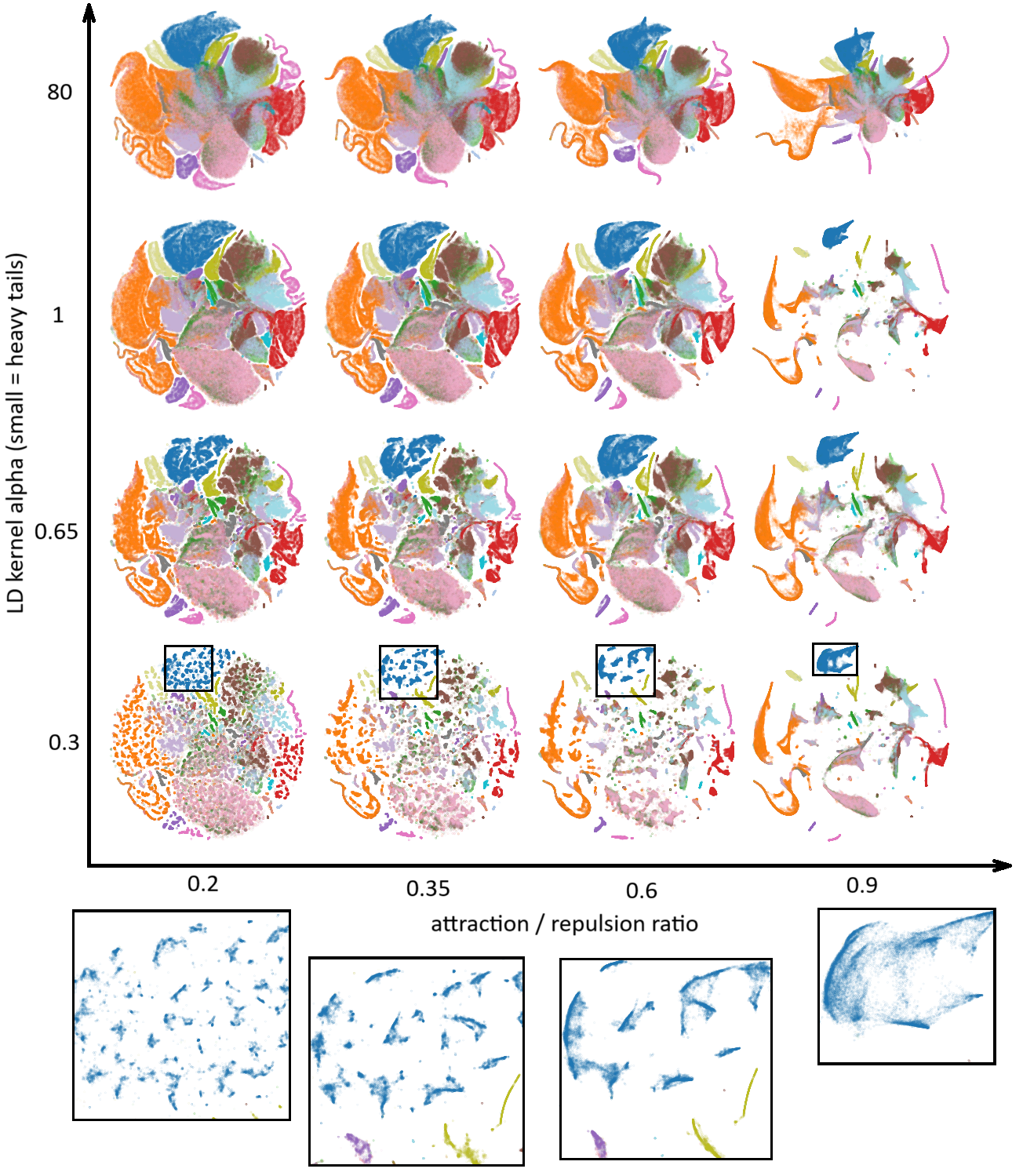}
        \caption{RNA-seq.: Tabula Muris Senis droplet}
        \label{fig:qualitative_bigmouse}
    \end{subfigure}
    \caption{Visual evaluation of the effect of the attraction/repulsion ratio and LD kernel tail thickness in embeddings produced using the proposed method.}
    \label{fig:qualitative_side_by_side}
\end{figure}

In the following paragraphs, the method is evaluated in the context of data visualisation by considering 2-dimensional embeddings. Two hyperparameters are central to the method: the attraction/repulsion ratio of the forces exerted on the data points, and the thickness of the tails in the LD kernels. Figure~\ref{fig:qualitative_side_by_side} shows the effect of these two hyperparameters visually on embeddings of single-cell datasets.\\

The left panel corresponds to `rat brain' \cite{Tasic}, where hundreds of messenger RNA sequences are measured in $23$ thousand cells found in the brains of rats. The colours were tailored by biologists to reflect the nature of the cells at hand, hot colours represent inhibitory neurons, cold colours correspond to excitatory neurons, and grey-ish colours are other types of cells present in the brain. In the right panel figures the `Tabula Muris Senis droplet' dataset \cite{TabulaMurisSenis}. This version of the dataset is comprised of a about $245$ thousand cells found in various tissues in mouse bodies, the points are coloured by tissue type.\\
In both datasets, increasing the thickness of the tails in the LD kernels (decreasing $\alpha$) fragments the data more and more, as observed previously on MNIST in Fig.~\ref{fig:taileffect}. When using the method for visualisation, heavy tails can produce LD clusters that are very dense, hampering visualisation. The attraction/repulsion ratio can be adjusted to prevent this collapse of points, and keep readable embeddings at low $\alpha$ values.\\
In unsupervised scenarios, the notion of cluster is ill-defined and subject to interpretation. This has been hinted in Fig.~\ref{fig:taileffect}, where looking at a dataset through different configurations of the proposed method can produce clusters of different granularity. If the proposed method is used in a continual optimisation scenario, where the user can modify the hyperparameter values on the fly and observe the resulting motion of points in real-time, the user can explore a broad band of the spectre of possible clusters by scanning through values of $\alpha$ and adjusting the attraction/repulsion occasionally. Capturing a hierarchy of different clustering solutions through multiple granularities is explored more thoroughly in the next subsection.\\

\begin{figure}[H]
    \centering
    \includegraphics[width=\linewidth]{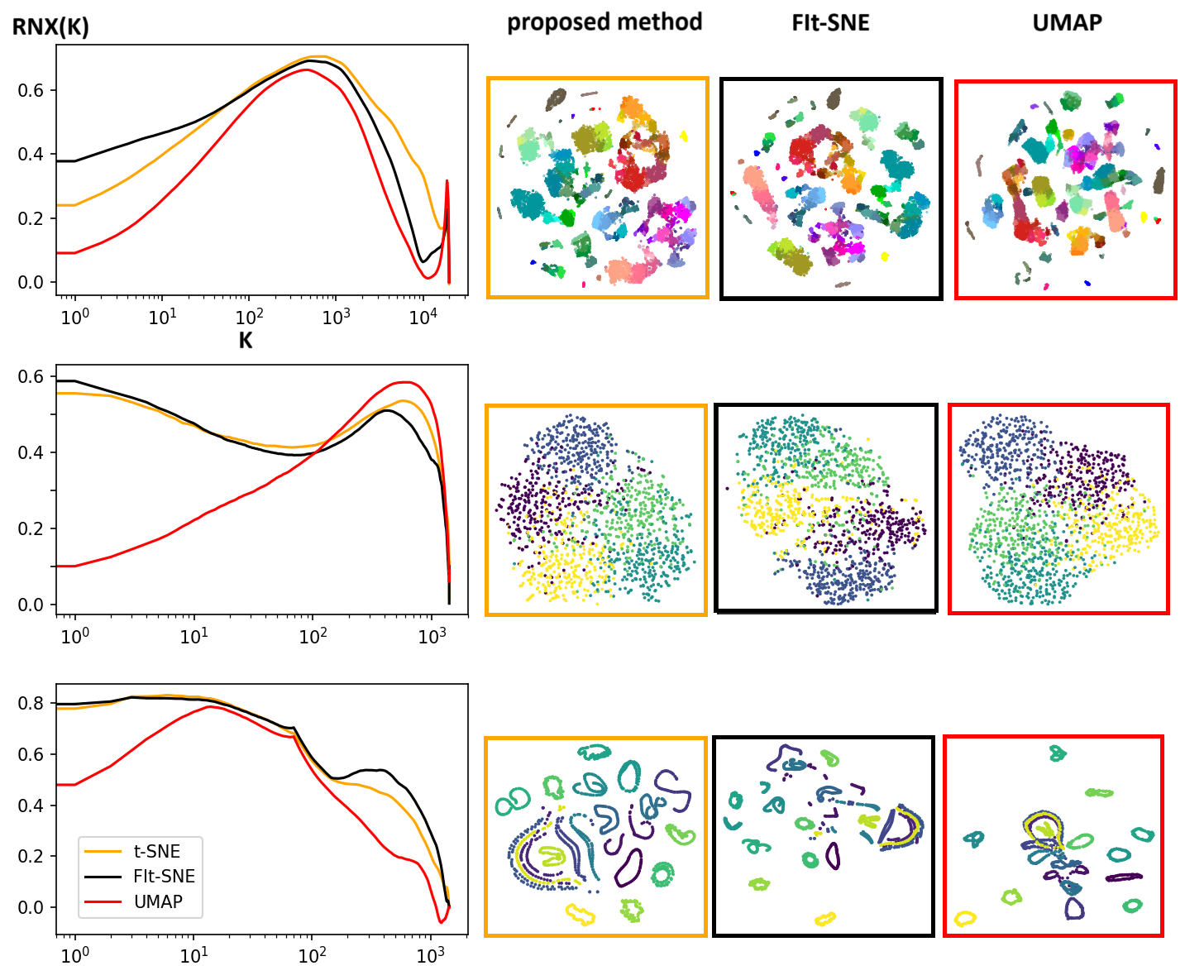}
    \caption{$R_\mathrm{NX}(K)$ curves of the proposed method, UMAP, and FIt-SNE on 3 datasets. From top to bottom: RNA-seq: rat brain, Gaussian blobs, COIL-20.}
    \label{fig:lesRNX}
\end{figure}

The embeddings produced in Fig.~\ref{fig:qualitative_side_by_side} appear sound when comparing them to other representations and explorations of the datasets \cite{artOfTsne, Tasic, TabulaMurisSenis}, but a more quantitative evaluation of the method is necessary to compare it to other modern NE algorithms. Figure ~\ref{fig:lesRNX} compares the proposed method to 2 staples of NE for visualisation: FI$t$-SNE \cite{fitsne} and UMAP \cite{umap}. These were chosen for their identical time complexity ($\mathcal{O}(N)$) and their wide adoption in practical NE. The 3 algorithms are compared across multiple scales on $3$ datasets using the RNX(K) \cite{RNX} quality metric. Values close to $1$ show good neighbourhood preservation at the given scale K, values close to $0$ indicates preservation equivalent to a random placement of the points. The rat brain dataset is figured in the top row, the middle row shows generated Gaussian blobs, and the $3^{\mathrm{rd}}$ row is COIL-20. The COIL-20 dataset consists of images of 20 objects rotating around an axis, drawing a ring like manifold in HD for each object. Default hyperparameters were used for FI$t$-SNE and UMAP, for the proposed method, the values for $\alpha$ and the attraction/repulsion ratio were chosen manually to produce embeddings that were visually similar to those obtained with FI$t$-SNE. \\
The quality curves show that the proposed method can preserve neighbourhoods across multiple scales similarly to FI$t$-SNE and UMAP. The systematic poor preservation of local structures in UMAP, consistent with the observations in \cite{hybrid, paperCyrilOverview} may be due to its reliance on unbiased negative sampling and on HD neighbours to determine the pairwise repulsive forces exerted on each point, allowing intrusive points in LD to remain undetected.

\begin{figure}[H]
    \centering
    \includegraphics[width=\linewidth]{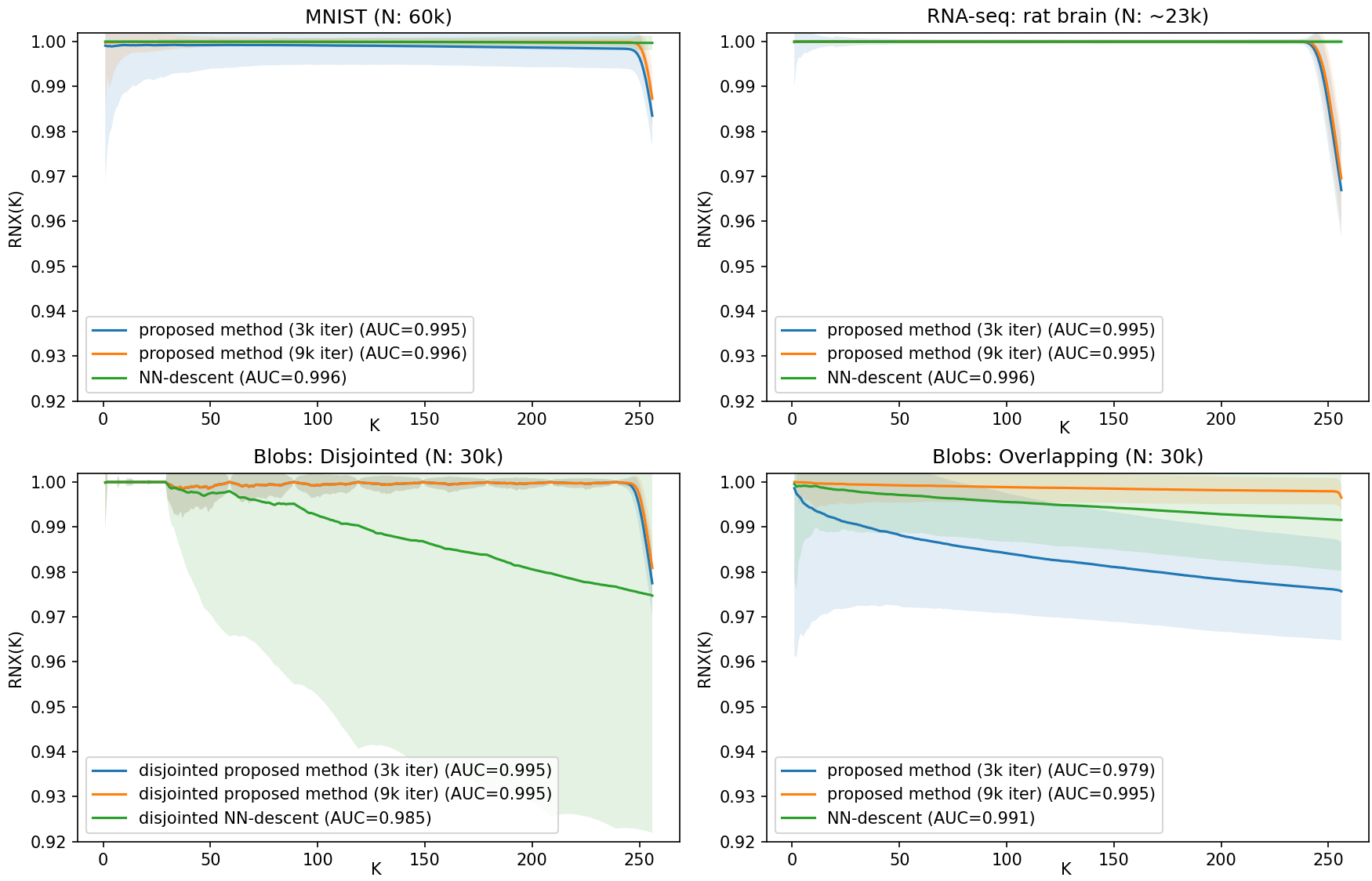}
    \caption{Comparison of the proposed neighbour finding algorithm with nearest neighbour descent. The x axis corresponds to the neighbourhood scale K, the y axis measures the quality of the estimated neighbourhood (higher is better).}
    \label{fig:KNN_vs_NNdescent}
\end{figure}

The proposed algorithm introduces a twist to nearest-neighbour descent \cite{NNdescent} by performing an iterative refinement of neighbours alongside the gradient descent steps producing the embeddings. When refining their neighbours, the two representations of the data communicate through their estimated KNN sets when producing candidates. Figure~\ref{fig:KNN_vs_NNdescent} compares the proposed KNN finding algorithm to nearest-neighbour descent on 4 datasets. The Y axis is the $R_\mathrm{NX}(K)$ criterion, which indicates the overlap between the estimated KNN sets and the ground truth exact sets: the closer to $1$, the closer the estimated sets are to the exact sets at scale K. The translucent bands indicate the standard deviation of $R_\mathrm{NX}(K)$ across the points: large bands indicate a high heterogeneity in the estimated KNN set quality across points. The sets that are evaluated here are the sets in the HD space.\\
Nearest-neighbour descent is a greedy refinement of neighbours which converges to a final solution, the proposed method requires an explicit termination criterion. For this experiment, the KNN refinement stops at the $3000^{\mathrm{th}}$ iteration for the blue lines, and $9000^{\mathrm{th}}$ iteration for the orange ones; the green lines are the RNX(K) curves for nearest-neighbour descent.\\
The two `Blobs' datasets are engineered to show two different scenarios: in the `Overlapping' case, the data is generated from $5$ Gaussian blobs with high overlap: the nearest neighbour descent manages to find good neighbours in its greedy approach. In the `Disjointed' case, there are $1000$ centres of $30$ points each, with much smaller standard deviations.
In the disjointed case, the isolation of the clusters in the HD space is a difficulty for nearest neighbour descent, whereas the proposed method managed to leap out of the local minimum and find better neighbour sets. Interestingly, the proposed method necessitated more time to reach close-to-perfect neighbour sets in the supposedly easier Overlapping case then in the Disjointed case.
The figure shows that the proposed nearest neighbour search can produce estimated sets that are competitive with those found by nearest-neighbour descent, given enough iterations.\\

\begin{figure}[H]
    \centering
    \includegraphics[width=\linewidth]{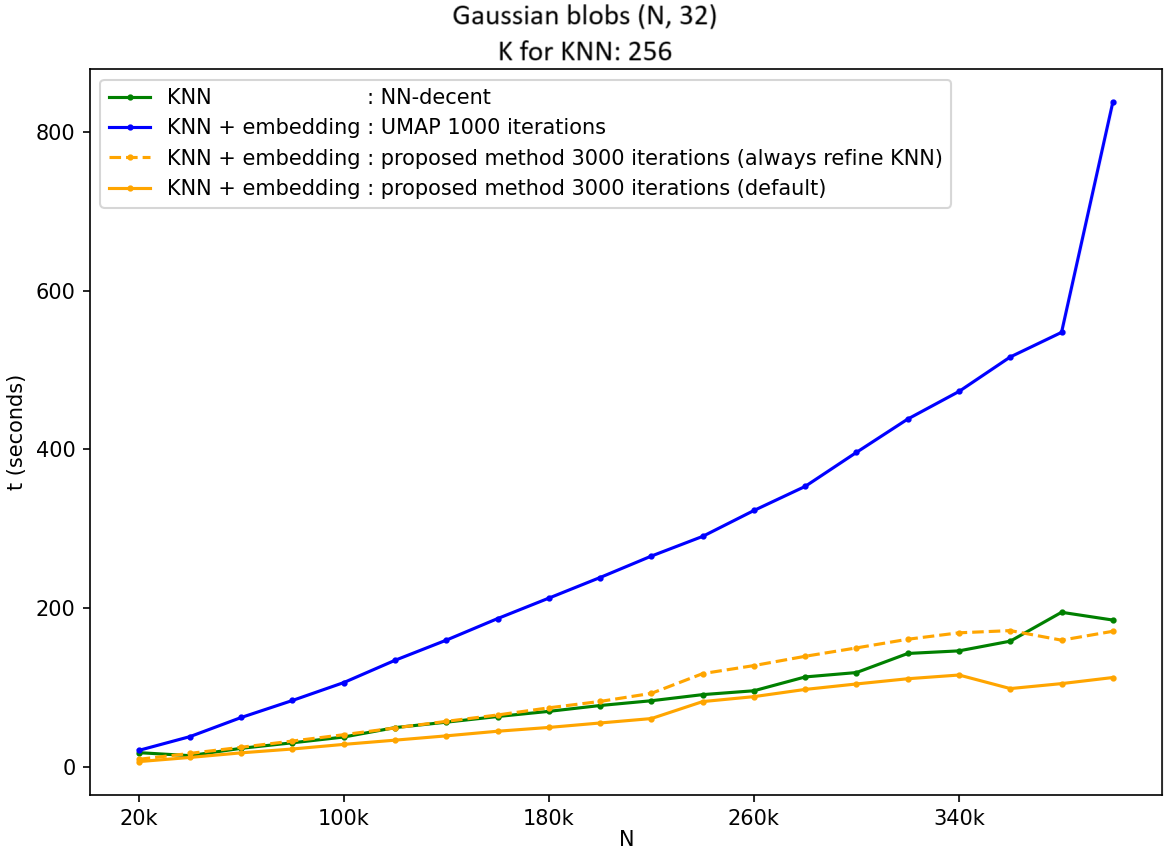}
    \caption{Effective time taken by the proposed method (laptop GPU), UMAP (laptop CPU), and nearest-neighbour descent (laptop CPU) on the same computer, across dataset sizes of fixed dimensionality (32).}
    \label{fig:time}
\end{figure}

As stated earlier, NE need to be correct, flexible, and fast. The previous figures show that the proposed method can produce competitive results and allows for flexibility in what structures to extract, by tweaking hyperparameters. A measure of the speed of the algorithm across different dataset sizes using a modern laptop is given in Fig.~\ref{fig:time}.\\

The yellow lines correspond to the proposed method, showing the time taken to perform $3000$ iterations. The dashed lines show the time taken if refining the HD neighbours at every iteration, the plain lines correspond to the default recommended configuration, where HD neighbours are refined following a probability depending on the estimated quality of the estimated sets, as explained in section $3$. The green line corresponds to the time taken by nearest neighbour descent, the blue one corresponds to $1000$ iterations of UMAP.\\

The linear time of the proposed method with respect to the number of points in the dataset is evident here. One cannot fairly compare the yellow lines to the blue and green lines because the proposed method runs on a GPU while the other methods in the experiment are not. In practice, the dimensionality of the HD space has a noticeable impact on the speed of the method. For the purpose of visualisation, it is recommended to reduce first the dataset dimensionality to manageable values by a linear projection, a standard practice in NE \cite{artOfTsne}. The sudden changes in the yellow and blue lines are likely specific to the hardware, where memory-related inefficiencies appear at certain dataset sizes.

\subsection{Exploring intermediate dimensionalities}

The proposed method is capable of producing embeddings of larger dimensionalities than 2 or 3, and can in theory adapt to new data points with little overhead. These two properties open the method to uses outside of the strict context of data visualisation. This section evaluates the method's capacity to extract relevant structures when representing data in more than $3$ dimensions through two use cases. In the first case, the algorithm is extended to allow a hierarchical representation of datasets, with computations take place in low but larger than 3 dimensions. The representations are evaluated qualitatively on two datasets. In the second scenario, the proposed method is used as a step to pre-process data in the difficult top-1 ImageNet classification task in a one-shot learning setting. Accuracy scores and visual inspection indicate that algorithm can produce embeddings that are well generalisable for a downstream supervised task.\\

In NE, the mismatch in the HD and LD kernel shapes can facilitate visualisation by exaggerating the separation between clusters in the LD space. The effect is visible in Fig.~\ref{fig:qualitative_side_by_side} and ~\ref{fig:taileffect}, where increasing the weight of kernel tails in LD increases the number of clusters that appear visually in the LD space. When using the method in an interactive environment, the user can get a sense of the hierarchical organisation of the data by observing how the clusters fragment and fusion depending on the changes in the hyperparameter $\alpha$. 
While useful as-is, the restrictive nature of very low dimensionalities can interfere with the optimisation and lead to results of heterogeneous quality, in particular if areas of the dataset exhibit high intrinsic dimensionalities. Here, an algorithm to capture the hierarchy through a graph is proposed, where the process of fragmenting clusters by lowering $\alpha$ is done in less restrictive dimensionalities.\\

Let us consider an embedding subject to a continual optimisation, where the LD kernel tails are slowly getting heavier in time. Snapshots of the changing embedding are taken at given intervals, the saved embeddings can be written $\mathcal{X}^\ell$, the level $\ell$ indexes the advancement in the growing kernel tail weight. A clustering algorithm is performed on each embedding $\mathcal{X}^\ell$ to extract the clusters $C^{(\ell)}$. A graph is then built to capture how the clusters evolved through time: each cluster is considered as a node, and the edges $e_{ij}$ between $C^{(g)}_i$ and $C^{(h)}_j$ are defined:

\[
e_{ij} = \beta \;.\;\frac{|\,C^{(g)}_i \:\cap\: C^{(h)}_j\,|}
   {\min\bigl(|C^{(g)}_i|,\;|C^{(h)}_j|\bigr)} \;, 
\quad \beta = 
\begin{cases}
 1 & \mathrm{if}\;|h - g| = 1,\\
 0 & \mathrm{else.}
\end{cases}
\]

\begin{figure}[H]
    \centering
    \includegraphics[width=\linewidth]{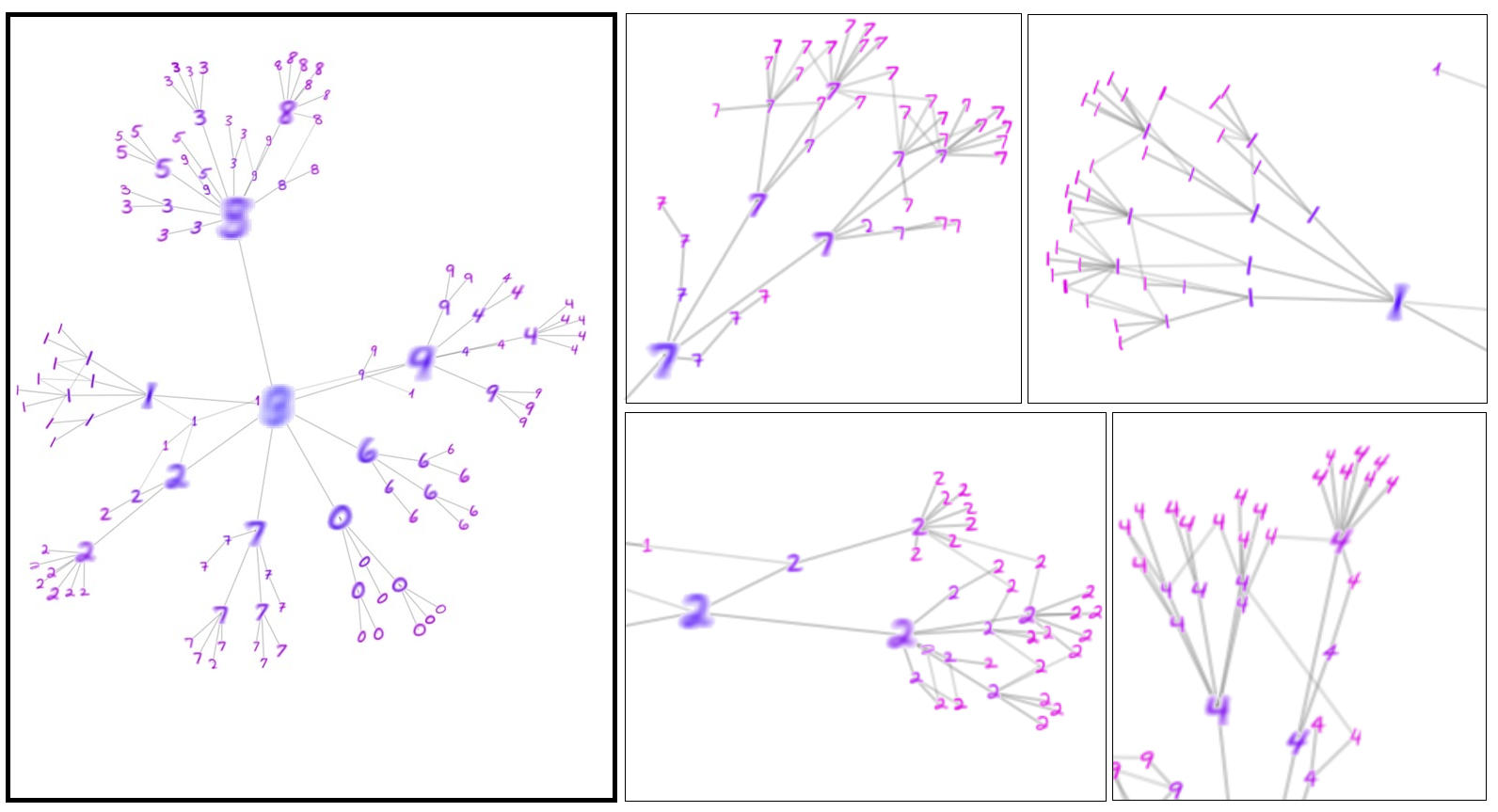}
    \caption{Hierarchical representation of MNIST, with LD dimensionality of $4$.}
    \label{fig:clustersMnist}
\end{figure}

\begin{figure}[H]
    \centering
    \includegraphics[width=\linewidth]{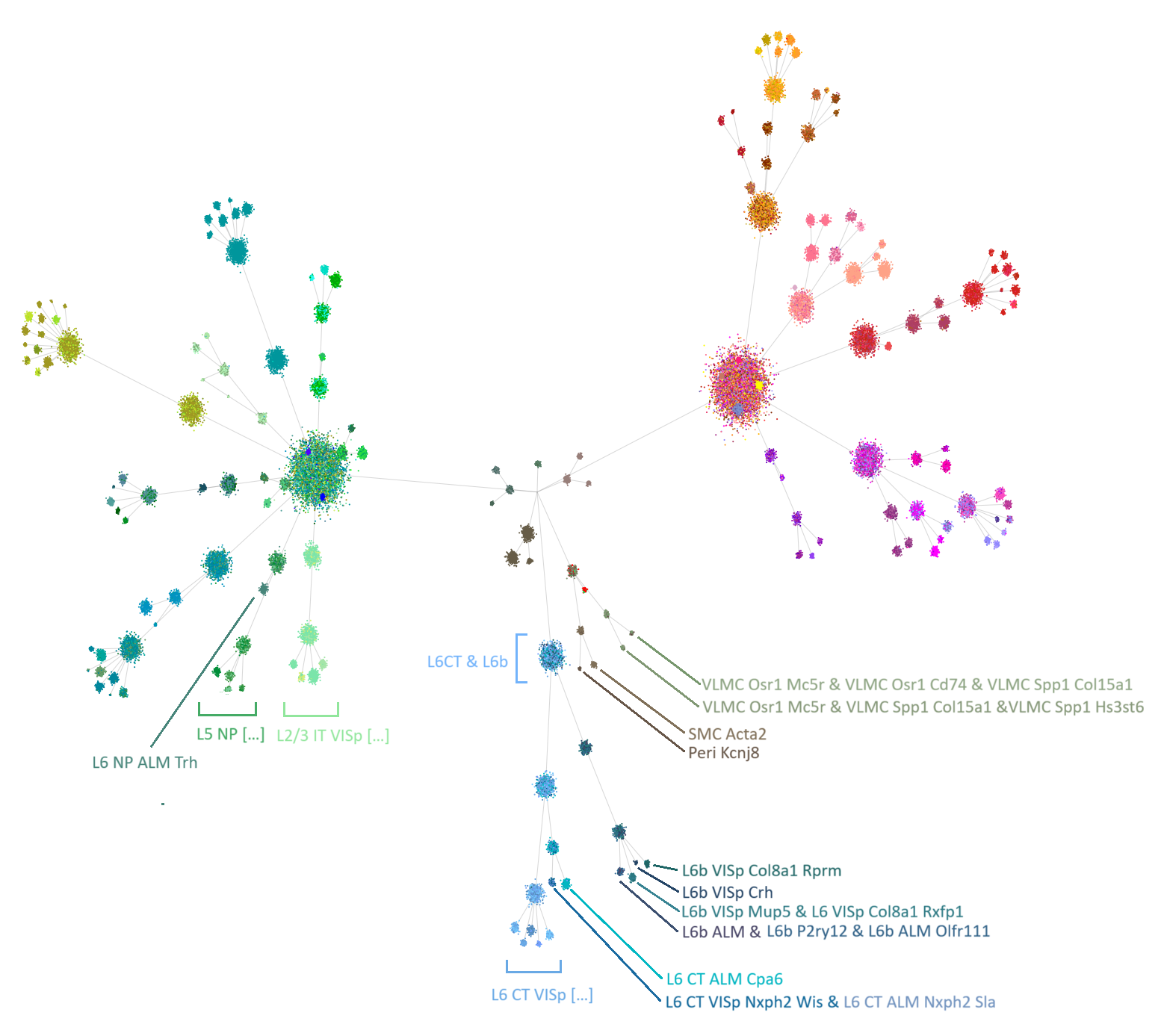}
    \caption{Hierarchical representation of the rat brain dataset, with LD dimensionality of $6$.}
    \label{fig:clsutersRat}
\end{figure}

Clustering is carried out here with DBSCAN \cite{dbscan}, chosen for its speed and ability to adapt to different number of clusters. In practice, the reasonably low embedding spaces and the tendency of NE to broaden gaps seem to facilitate the work of DBSCAN in this experiment. Once the graph built, it can be used algorithmically or visualised using a force-directed layout, as in Figures ~\ref{fig:clustersMnist} and ~\ref{fig:clsutersRat}. In both figures, a central node representing the whole dataset is added for aesthetic reasons, and the size of the nodes is proportional to $\sqrt{|C^{\ell}_i|}$, with bounds.\\ 

In Figure ~\ref{fig:clustersMnist}, the nodes are represented by their mean image in HD. The view on the left represent the whole dataset using three levels, the views on the right show close ups if performing the same procedure with an additional level of LD kernel heaviness, revealing finer-grained details.
A connoisseur in the art of writing down numbers by hand might identify a meaningful organisation in these representations. For instance, in the view on the left, the $5$, $3$, and $8$ digits speciate a level of depth later than the rest, and indeed they differ only by small segments. In the same view, the rare images of $1$ that exhibit a bottom horizontal line are represented here as a mixture between $1$ and $2$, and indeed $2$ is the only other digit that can have a straight horizontal line at the bottom. The finer grained organisation shown on the right also reveals subjectively meaningful structure upon scrutiny.\\

The hierarchical structure of the rat brain dataset \cite{ratbrain} is also extracted, the results are shown in Fig.~\ref{fig:clsutersRat}. In this figure, instead of showing the depiction of a prototype for each node, the points belonging to each cluster are rendered as Gaussian noise using the colours that were hand-picked by the authors of \cite{ratbrain}. A hierarchy of clusters is apparent and bears a strong resemblance with the ground-truth dendrogram accompanying the dataset in original paper, which indicates the relationship between the cells as currently established in biology.\\

It is difficult to define what a cluster is, in practice the definition is often tailored to a task or interpreted by a human. These experiments show that the capacity to control with which severity a NE will tear manifolds in their LD representations can help in the extraction of meaningful clusters across selected scales. The NE performs what one might describe as a tweakable "pre clustering" of the data, facilitating and directing the work of a following conventional clustering algorithm.\\

To further evaluate the soundness of higher dimensional NE of data, the rest of the section will consider a 32-dimensional NE of the ImageNet dataset obtained using the proposed method on the first $192$ principal components of the $1280$-dimensional latent representation of the images, as extracted by the vision encoder of the large vision-language model EVA \cite{EVA} (images $\rightarrow$ $1280$-dim. latent space $\rightarrow$ $192$ PCs $\rightarrow$ $32$-dim. NE). The first $192$ principal components capture $71.1\%$ of the variance of the $1280$-dimensional dataset. The preliminary linear projection towards $192$ dimensions is performed to reduce the runtime of the NE, which scales linearly with the area of the data matrix. Furthermore, the current implementation on GPU has a hardcoded limit on the dimensionality of the input data, this limit was set to facilitate development while keeping reasonably optimised memory access patterns and GPU kernel launch overheads.
The ImageNet dataset is comprised of more than $1.2$M natural images of $1000$ distinct classes. The number of classes makes it a particularly challenging benchmark, especially in weakly supervised scenarios.\\
In order to evaluate the embedding, a simple $1$-nearest neighbour classifier using Euclidean distances is trained in different learning contexts and on different representations of ImageNet, the accuracies are presented in Table ~\ref{table}. In the one-shot leaning scenario, the mean accuracy of $100$ trials is shown; a trial consist of revealing the label of a random point of each class, and classifying all the other points.\\

\begin{table}[H]
    \centering
    \small
    \begin{tabular}{ccccc}
         & 1280, EVA & 192, PCA & 32, NE\\
       one-shot (top-1)  & 47.3\% & 45.9\% & \textbf{76.2\%}\\
       one-shot (top-5)  & 67.3\% & 67.8\% & \textbf{92.0\%}\\
       crossval (top-1, train)  & 84.9\% & 83.2\% & \textbf{87.2\%}\\
       crossval (top-1, test)  & 87.2\% & 87.2\% & \textbf{88.7\%}\\
    \end{tabular}
    \caption{Comparison of classification performance in $3$ representations of the ~$1.2$M images of ImageNet, in 1 shot learning and 10-fold cross-validation.}
    \label{table}
\end{table}

The performance in one shot learning and the tighter margin between train and test accuracies in cross-validation indicate that, despite being obtained in a unsupervised manner, the $32$-dimensional NE exposes information in a way that facilitates generalisation regarding this supervised task. The embedding is publicly available at \textit{https://github.com/PierreLambert3/Imagenet-tSNE-embedding.git}.\\

\begin{figure}[H]
    \centering
    \includegraphics[width=\linewidth]{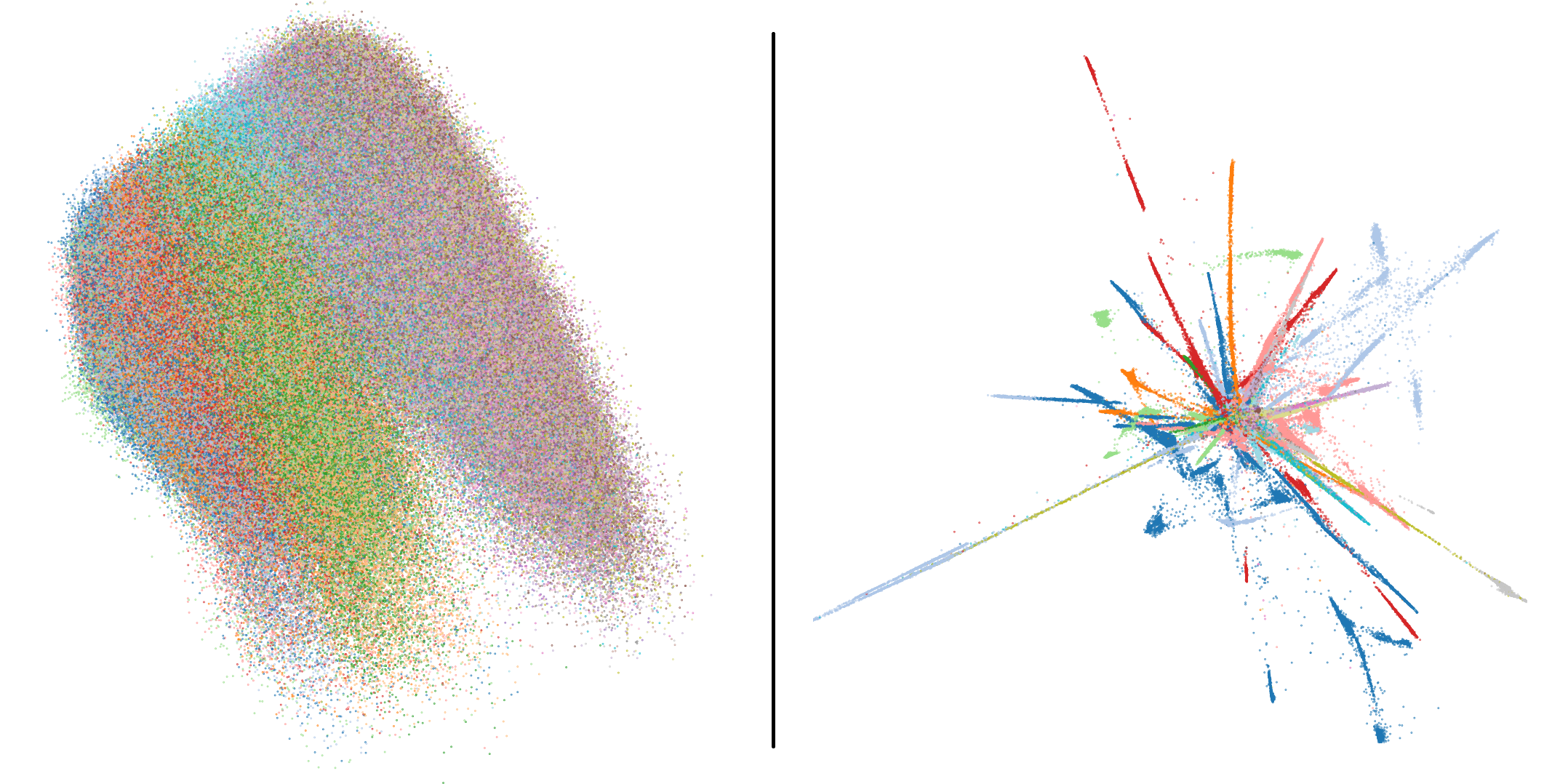}
    \caption{Two-dimensional PCA projections of the 1280-dimensional representation of ImageNet by EVA. Left: raw EVA features (1280 (EVA) $\rightarrow$ 2 (PCA vis.)); right: EVA features after embedding in 32D with the proposed method (1280 (EVA) $\rightarrow$ 192 (PCA) $\rightarrow$ 32 (t-SNE) $\rightarrow$ 2 (PCA vis.)).}
    \label{fig:effectOfTSNE}
\end{figure}

To shed some light on how the NE organises data in the $32$-dimensional space, Figure~\ref{fig:effectOfTSNE} shows a $2$-dimensional PCA projection of the embedding, along with a PCA projection of the initial EVA representation. The linear nature of PCA allows an incomplete but undistorted view of the structures. The NE seems to represent data in tighter, less diffuse groups of points than the original representation. An artifact common to PCA and to spectral clustering is apparent \cite{laplacSpectral, attracRepulSpectrum}, where only a few clusters are distinct and appearing like long spikes, while the others collapse towards the centre of the linear projection.

\section{Conclusion}

This paper shows that good quality in neighbour embedding  can be attained quickly and without upper limit on the dimensionality of the LD space, by combining the tracking of neighbours in the LD space with negative sampling. The proposed method relies on a novel extension of nearest neighbour descent, where the sets of neighbour around each point get refined iteratively concurrently with the embedding; this process benefits from the embedding in construction to refine neighbour sets faster, as a source of neighbour candidates that are likely to be eligible by design and often confirmed.\\
The method could supposedly be applicable to dynamical contexts with streams of data that are not known in advance, or data points with dynamical values. The virtually non-existent overhead of adding, removing, or modifying the data in the HD side results from the fact that HD similarities and neighbourhoods are estimated alongside the embedding, with a heuristic allowing to divert compute to the HD side or the embedding dynamically, depending on the need. This contrasts with other methods that typically run through several distinct and different algorithmic stages, like data preparation (e.g., conversion of the HD coordinates into a sparse neighbourhood graph, pre-computation of entropic affinities, etc.) before actually launching the NE iterations. This could broaden the applicability of NE to visualisation in continual learning, but also to contexts outside of strict visualisation for instance to pre-process data before some other computational tasks, like clustering or classification.\\ 



\begin{footnotesize}

\bibliographystyle{unsrt}
\bibliography{bibli}

\end{footnotesize}

\end{document}